%% file: witte2025wacv.tex
\documentclass[10pt,twocolumn,letterpaper]{article}

\usepackage[pagenumbers]{wacv} 
\usepackage{graphicx}
\usepackage{amsmath}
\usepackage{amssymb}
\usepackage{booktabs}
\usepackage{flushend}

\input{stachnisslab-latex}

\usepackage[pagebackref,breaklinks,colorlinks]{hyperref}
\usepackage{multirow}
\usepackage{booktabs}
\usepackage{svg}
\usepackage{array}
\usepackage{booktabs}

\usepackage[capitalize]{cleveref}
\crefname{section}{Sec.}{Secs.}
\Crefname{section}{Section}{Sections}
\Crefname{table}{Table}{Tables}
\crefname{table}{Tab.}{Tabs.}

\def\wacvPaperID{884} 
\def\confName{WACV}
\def\confYear{2025}

\newcommand{\name}{EAFormer}

\begin{document}

\title{Epipolar Attention Field Transformers \\for Bird's Eye View Semantic Segmentation}

\author{
Christian Witte$^{1,2}$
\and
Jens Behley$^2$
\and
Cyrill Stachniss$^{2,3}$
\and
Marvin Raaijmakers$^1$ 
\vspace{0.05cm} 
\and  
$^{1}$CARIAD SE, Germany 
\and 
$^{2}$Center for Robotics, University of Bonn, Germany 
\and 
$^{3}$Lamarr Institute for Machine Learning and Artificial Intelligence, Germany
}

\maketitle

\begin{abstract}
  Spatial understanding of the semantics of the surroundings is a key capability needed by autonomous cars to enable safe driving decisions.
  Recently, purely vision-based solutions have gained increasing research interest. In particular, approaches extracting a bird's eye view~(BEV) from multiple cameras have demonstrated great performance for spatial understanding. 
  This paper addresses the dependency on learned positional encodings to correlate image and BEV feature map elements for transformer-based methods.
  We propose leveraging epipolar geometric constraints to model the relationship between cameras and the BEV by Epipolar Attention Fields. They are incorporated into the attention mechanism as a novel attribution term, serving as an alternative to learned positional encodings.
  Experiments show that our method \name\ outperforms previous BEV approaches by 2\% mIoU for map semantic segmentation and exhibits superior generalization capabilities compared to implicitly learning the camera configuration.
\end{abstract}

\section{Introduction}
\label{sec:intro}

\begin{figure}[tb]
  \centering
  \includegraphics[height=7.5cm]{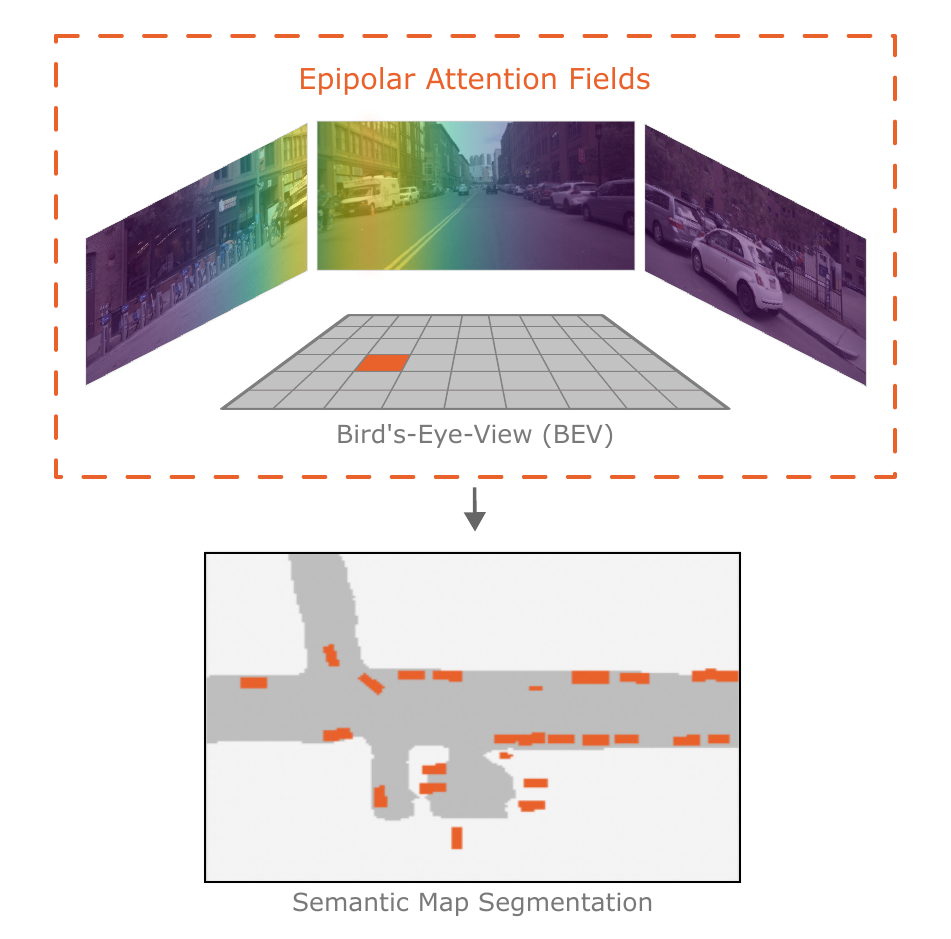}
  \caption{Our approach attends features of multi-view camera images to  BEV features by computing Epipolar Attention Fields (top) instead of leveraging learnable positional encoding for BEV semantic segmentation (bottom), where we predict the drivable area (dark grey) and vehicles (orange).}
  \vspace{-0.2cm}
  \label{fig:motivation}
\end{figure}


Robust perception of the surrounding environment is a key component for enabling autonomous driving. Most autonomous systems heavily rely on multi-modal sensor setups to ensure the high safety demands. Recently, pure camera-based perception for autonomous vehicles has been a prominent research topic.
Bird's eye view~(BEV) transformation methods using multiple cameras allow for a versatile representation that can be used for various downstream tasks such as map semantic segmentation~\cite{peng2023wacv,yang2023cvpr,yuan2024wacv,chen2022arxiv,zhou2022cvpr} or 3D bounding box detection~\cite{xie2022arxiv, yang2023iccv,huang2021arxiv-bevdet}. 
The objective of using BEV is to create a coherent and semantically meaningful top-view representation of the environment by fusing sensor features into a single representation with a shared coordinate system~\cite{zhang2022arxiv-beverse}. 

One line of work for camera-only BEV perception directly projects~\cite{philion2020eccv} or samples~\cite{harley2023icra} image features with the known camera parameters. 
In contrast, transformer-based BEV projection approaches like CVT~\cite{zhou2022cvpr} and GKT~\cite{chen2022arxiv} learn the relationship between image features and BEV grid elements by implicitly encoding camera parameters into the positional embeddings. 

For our approach, we propose to consider the BEV feature plane as an additional camera view and we reformulate the mapping between image and BEV features as a multi-view problem. A point projected onto the BEV grid plane then corresponds to an epipolar line in the camera views. 
We extend this explicit geometric relationship to epipolar fields, which are a normal distribution imposed on the distance to the epipolar line. The epipolar fields allow us to capture important neighboring features and to account for the cell size of the BEV grid. We use the epipolar fields to weigh the attention of the transformer and thus enable explicit geometric reasoning in the attention mechanism. This eliminates the need for learning the geometric correlation between BEV and image features as positional encodings. Our approach is illustrated in \cref{fig:motivation}.

As we do not rely on a learnable positional encoding, our method can more easily adapt to new camera setups and is more robust against changes in the camera parameters. This is advantageous as each vehicle model has a slightly different camera setup. It is impracticable to retrain the entire network from scratch but rather to fine-tune on each camera setting. Experiments demonstrate that our approach significantly outperforms implicit positional encoding for zero-shot transfer to a completely different camera setting.
Furthermore, the best-performing algorithms on recent datasets do not necessarily generalize very well to semantic map segmentation. As our experiments suggest, our approach outperforms other transformer-based methods and shows a strong generalization capability.

To summarize, our contributions are twofold:
\begin{itemize}
\setlength{\itemsep}{0pt}
\setlength{\parskip}{0pt}
    \item We leverage explicit geometric reasoning for transformer-based BEV projection using Epipolar Attention Fields~(EAFs), eliminating the need for positional encodings. We integrate this idea into a BEV semantic segmentation network called~\name.
    \item Experiments show that our approach outperforms state-of-the-art approaches for semantic BEV prediction and demonstrates strong generalization capabilities on data collected with different camera setups.
\end{itemize}

\section{Related Work}
\label{sec:related-work}

\subsection{BEV-based Semantic Segmentation}
Recent work on multi-view camera perception focuses on projecting image features into a BEV representation, which has been employed for different downstream tasks such as map-based semantic segmentation.

To obtain a BEV representation, OFT~\cite{roddick2018arxiv} samples image features for predefined 3D voxels and projects the voxel volume by an orthographic transformation onto a 2D ground feature plane. Similarly, Simple-BEV~\cite{harley2023icra} bi-linearly samples image features for each voxel in an efficient manner. 
Lift-Splat-Shoot (LSS)~\cite{philion2020eccv} and derivative work~\cite{huang2021arxiv-bevdet, li2023aaai, liu2023icra, zhou2023cvpr} forward-project image features onto a voxel volume using a categorical depth estimation and collapse the accumulated features onto the 2D feature plane by an efficient pooling operation. 
M$^2$BEV~\cite{xie2022arxiv} utilizes a uniform depth distribution and incorporates a joint training for 3D object detection.  Further work extends the idea of LSS to multiple timesteps and applies the view transformation to perform motion forecasting \cite{hu2021iccv, zhang2022arxiv-beverse}, incorporate parametric depth distribution \cite{yang2023iccv} or employ residual graph convolutional networks \cite{chen2024wacv}. 
Although depth uncertainty is accounted for in the 3D-to-2D projection, forward-projecting methods depend on the accuracy of depth estimation. 

Other approaches project image features onto a BEV feature grid by computing attention~\cite{vaswani2017neurips} between BEV queries and image features. To control the quadratic computation demands, Saha~\etal\cite{saha2022icra} attend the BEV queries to vertical scanlines of the images. BEVFormer~\cite{li2022eccv, yang2023cvpr} leverages deformable attention to sparsely cross-attend BEV and image features, while using past BEV queries for temporal fusion.

Cross-View Transformers (CVT)~\cite{zhou2022cvpr} utilize a geometry-aware positional encoding to learn a correspondence between camera coordinates and BEV grid locations by computing cross-attention between BEV queries and image features directly. Geometric-guided Kernel Transformer (GKT)~\cite{chen2022arxiv} limits the cross-attention computation via a kernel function using the 2D projection as prior. CoBEVT~\cite{xu2022arxiv} extends the idea of CVT to cooperative multi-agent map prediction and introduces a hierarchical structure and a fused axial attention module.
BAEFormer~\cite{pan2023cvpr} leverages early fusion of image and BEV features by a bi-directional cross-attention mechanism. For each hierarchical image feature map, the authors cross-attend BEV queries with image features and, vice versa, image queries with BEV embeddings. 
Although PETR~\cite{liu2022eccv} employs instance queries in contrast to BEV queries like CVT~\cite{zhou2022cvpr}, it is important to note that PETR and other approaches based on PETR~\cite{liu2023iccv, yan2023iccv} also implicitly encode the camera parameters into the positional encoding. This is achieved by sampling a fixed number of 3D points along the viewing ray for each image feature and feeding these points to an MLP to compute the feature's positional encoding.

The idea of multi-view BEV transformation has been further extended to a multi-modal setting \cite{liu2023icra, salazar2023arxiv, zou2023arxiv, wang2023iccv, yan2023iccv} as well as to multiple time steps toward temporal fusion~\cite{han2024ral, qin2023cvpr}.
In summary, attention-based methods aim to leverage the superior modeling performance of transformers. 
However, geometric relationships are often only implicitly encoded, potentially leading to overfitting to the camera setups. In contrast, this paper presents an approach that explicitly encodes geometric relationships between the image and BEV features.

Therefore, we combine the advantages of transformer-based approaches with explicit geometric reasoning.

\subsection{Epipolar Transformers}
He \etal~\cite{he2020cvpr-epipolar} leverage epipolar geometry in combination with attention for multi-view 3D human pose and joint estimation. Given a 2D point $p$ in a reference view, the aim is to find the corresponding point $p'$ in a neighboring view. The matching is obtained by computing attention weights between the point $p$ and sampled points $p_i'$ along the projected epipolar line in the neighboring view. This is densely repeated for all points $p_j$ in the reference view.
As this work only captures semantic information strictly along the epipolar line, TransFusion~\cite{ma2021bmvc} introduces the notion of epipolar fields expressing a similarity based on the distance to the epipolar line. 

To the best of our knowledge, this paper is the first one to leverage epipolar fields as an alternative to positional encoding and to exploit epipolar geometry for BEV perception by guiding the attention with Epipolar Attention~Fields.

\section{Method}
The goal of our multi-view BEV approach is to correlate image features with BEV features to learn a versatile and efficient representation for map-based semantic segmentation. As for CVT~\cite{zhou2022cvpr}, we obtain image feature maps for each camera and model the correspondence to BEV features by iteratively aggregating image features using the cross-attention mechanism. Instead of a learnable positional encoding, we employ Epipolar Attention Fields to explicitly model geometric constraints and thus the spatial correlation.

To illustrate our idea, we will first introduce the multi-view formulation for BEV settings in \cref{sec:multi-view}. Then, we present our concept of attention weighting to attribute for known relationships. As our main contribution for explicit geometric modeling in transformers, we introduce Epipolar Attention Fields in \cref{sec:epipolar-attn}. We propose an architecture for BEV-based semantic segmentation based on Epipolar Attention Fields, which we termed \textbf{E}pipolar \textbf{A}ttention \textbf{F}ield Transf\textbf{ormer}s (\textbf{\name}), and which we explain in detail in~\cref{sec:arch}. 

\subsection{Multi-View Formulation}
\label{sec:multi-view}
\begin{figure}[tb]
  \centering
  \includegraphics[width=\linewidth]{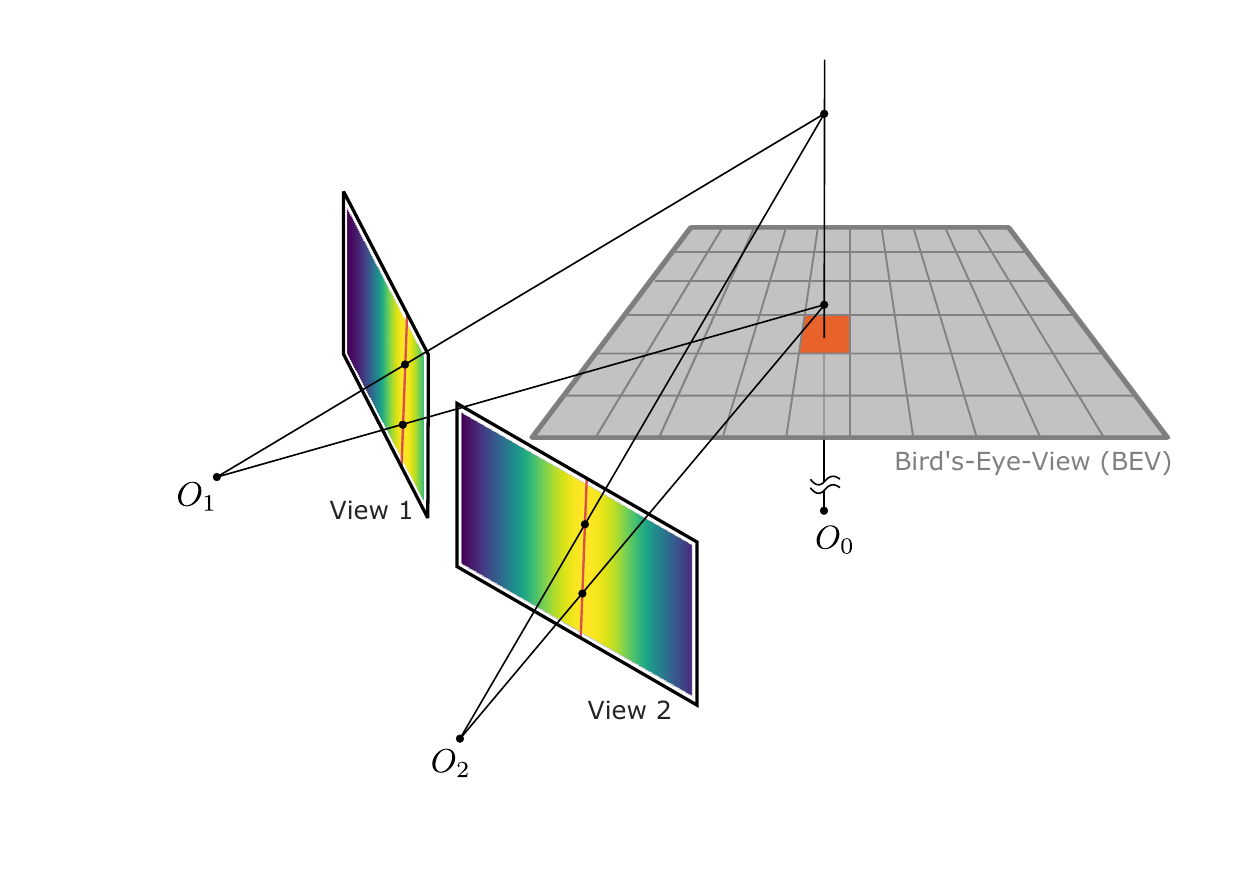}
  \caption{The BEV (grey) is considered the reference view for epipolar geometry. We project each coordinate of the grid cell center onto all other views. A given grid cell (red) thus corresponds to the epipolar lines projected onto the source views 1 and~2. We impose a normal distribution on the distance to the epipolar line and obtain Epipolar Attention Fields, which are visualized as heatmaps in the source views.}
  \label{fig:multi-view}
\end{figure}
As our approach utilizes epipolar geometry~\cite{hartley2004book} for attention weighting, we first provide a brief overview following the notation introduced by Hartley and Zisserman~\cite{hartley2004book}.
We denote the 2D BEV grid as an image plane of the $0$-th view with the projection center $O_0$, which is infinitely far away with parallel rays passing through the image plane. The multi-view images are captured by $N$ cameras, each with a projection center $O_n, n\in\{1, \dots, N\}$.

Let $\mathbf{x}$ be a homogeneous representation of a 3D point and $\mathbf{x}_0$ be its 2D orthogonal projection onto BEV plane respresented by homogeneous coordinates.
Given the height/depth ambiguity of this multi-view setting, all points of the equivalence class of $\mathbf{x}$, \ie, all points along the ray on $(O_0, \mathbf{x})$, correspond to a line in the $i$-th image and can be modeled by the essential matrix $E_i$. A point $\mathbf{x}_i$ in the $i$-th view maps to point $\mathbf{x}_0$ if
\begin{align}
    {\mathbf{x}_i}^\intercal  E_i \mathbf{x}_0 = 0
    \label{eq:epipolar}
\end{align}
with $\mathbf{l}_i = E_i \mathbf{x}_0$ being the epipolar line for the $i$-th view and point $\mathbf{x}_0$.

Thus, all points in 3D space that correspond to the point $\mathbf{x}_0$ in the BEV plane, must project to a point on the epipolar line~$\mathbf{l}_i$ in view $i$.
\cref{fig:multi-view} visualizes the multi-view formulation.

\subsection{Attention Weighting}
\label{sec:attn-weighting}
As introduced by Vaswani \etal~\cite{vaswani2017neurips}, the attention mechanism can be trained to learn the dependencies between queries and keys/values.
Since the attention computation between queries $Q$ and keys $K$ is permutation equivariant, the position and sequence information is commonly encoded as positional encodings, which are added to the keys and queries.

These positional encodings serve the purpose of learning how positions of features correlate to each other yielding high attention values for high correlation and accordingly low attention values for little correlation.
For BEV perception, physical or a priori knowledge about the correlation between values/keys and queries can be incorporated into the positional encoding implicitly \cite{zhou2022cvpr}. 

We propose to eliminate the necessity of positional encodings for our multi-view setting and introduce attention weighting as an alternative approach to explicitly ingest known correspondence of values/keys and queries. 

To this end, we introduce attention weights for the attention computation.
We define $Q \in \mathbb{R}^{n_q \times d}$ and $K, V \in \mathbb{R}^{n_k \times d}$, where $n_q$ is the number of queries, $n_k$ is the number of keys (which is equal to the number of values) and $d$ is the size of the keys, values and queries. Furthermore, we define $W \in \mathbb{R}^{n_q \times n_k}$, of which each element is the weight for a query-key pair.

Then, we adapt the original attention function~\cite{vaswani2017neurips} by incorporating the weights of $W$ resulting in:
\begin{align}
    Attention(W, Q, K, V) &= \mathit{softmax}\left(W \odot \frac{QK^\intercal}{\sqrt{d_k}}\right) V
\end{align}
where $\odot$ is the Hadamard product.
Rather than adding positional encodings to the keys $K$ and queries $Q$, we encode the correlation between a key-query pair through the attention weights in matrix $W$. 
Since a key corresponds to a position in a camera image and a query corresponds to a position on the BEV grid, we can use epipolar geometry to express how well a key-query pair matches geometrically.

\subsection{Epipolar Attention Fields}
\label{sec:epipolar-attn}

CVT~\cite{zhou2022cvpr} learns the matching between the keys and queries by encoding the geometric information (\ie, extrinsics and intrinsics) into positional embeddings. 
However, with multi-view geometry, we know an accurate correlation between features in different views, as described in \cref{sec:multi-view}. Features in one view must lie on or near the epipolar line in the other view. To exploit this prior knowledge, we model the geometric relationship with Epipolar Attention Fields, which we use for attention weighting.

In this section, we describe how to compute the attention weight matrix $W$ based on Epipolar Attention Fields.

Consider the $q$'th query in $Q$ and the $k$'th key in $K$. The element $W_{q,k}$ in matrix $W$ represents the attention weight for this query-key pair. 
Let $i$ denote the view from which key $k$ originates and $\mathbf{x}_i$ denote the image feature position (in homogeneous coordinates) corresponding to $k$. Given the essential matrix $E_i$, we project the BEV position corresponding to query $q$ onto the $i$-th view obtaining the epipolar line $\mathbf{l}_{i}$. 

Ideally, a point in the $i$-th view must be exactly on the epipolar line of a point from the BEV view if those two points correspond to each other (\ie, \cref{eq:epipolar} must hold).
However, similar to Epipolar Fields~\cite{ma2021bmvc}, we observe that important features do not necessarily lie on the epipolar line but in its vicinity.
We, therefore, model the attention weight $W_{q,k}$ as a Gaussian over the distance between point~$\mathbf{x}_i$ and epipolar line $\mathbf{l}_i$. In this way, we impose high attention weights for $\mathbf{x}_i$ near the epipolar line and low attention weights for elements farther away.
The distance between point~$\mathbf{x}_i$ and line $\mathbf{l}_i$ can be computed by taking the dot product of $\mathbf{x}_i$ and $\mathbf{\hat{l}}_i$, where $\mathbf{\hat{l}}_i$ is $\mathbf{l}_i$ after applying normalization for homogeneous lines on it.

Based on the observations above, we propose the following equation for the attention weight: 
\begin{align}
    W_{q,k} &= \exp\left( - (\lambda \lambda_{q,i})^2 \left( \mathbf{x}_i \mathbf{\hat{l}}_{i}^\intercal \right)^2 \right)
\end{align}
where $\lambda$ and $\lambda_{q,i}$ determine the width of the Gaussian distribution. The former, $\lambda$, is a hyperparameter, which we call the distance-strength parameter.

\begin{figure}[tb]
  \centering
  \includegraphics[width=\linewidth]{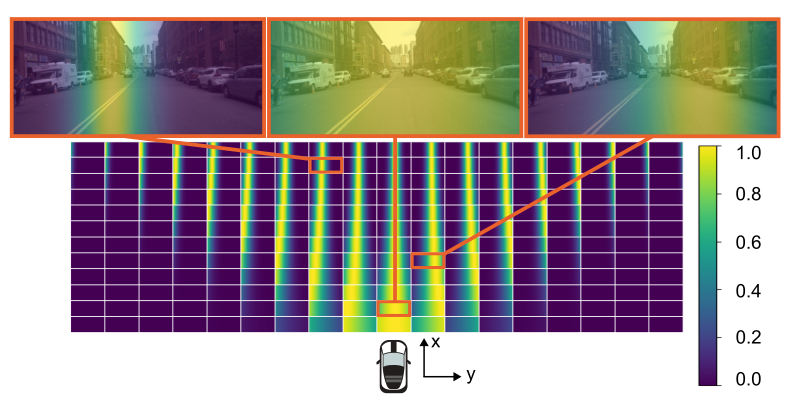}
  \caption{Epipolar Attention Fields for the front camera of nuScenes. The grid (middle) visualizes the BEV cells that are visible to the front view camera. The imaginary location of the vehicle is at the bottom center of the grid, facing upward.
  For illustrative purposes, each BEV cell is visualized by the Epipolar Attention Field heatmap of a corresponding query. Yellow indicates high attention weights, while blue indicates low attention weights. For three distinct BEV grid cells (red outline), we overlay the heatmap with the actual input image showing the Epipolar Attention Field.}
  \label{fig:epipolar-fields-front-view}
  \vspace{-0.2cm}
\end{figure}

For the BEV cell corresponding to $q$ and an image feature map $\mathcal{F}$, we define the \emph{Epipolar Attention Field} (EAF) to be the set of $W_{q,k}$ weights for all $k$'s that correspond to a feature in $\mathcal{F}$.
\cref{fig:multi-view} visualizes the EAFs for a given BEV cell and two different views/feature maps.

\begin{figure*}
  \centering
  \includegraphics[height=7cm]{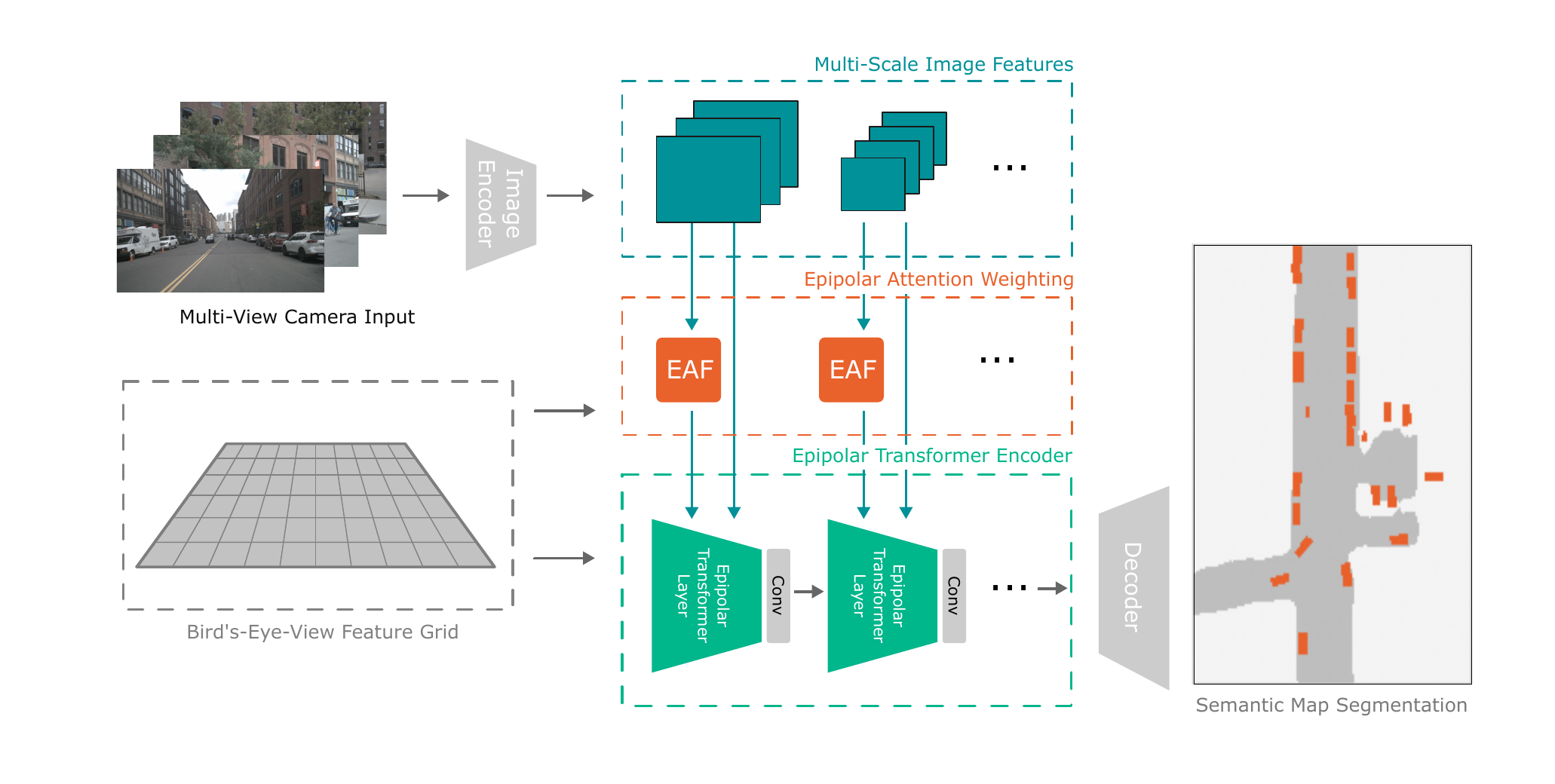}
  \caption{An illustration of our \name \ architecture. For each input image, we obtain multi-scale image feature maps. We compute the Epipolar Attention Fields~(EAF) based on the image feature coordinates, the BEV feature grid coordinates, and the camera parameters. The Epipolar Transformer Encoder attends the BEV grid queries to the image feature iteratively over the multi-scale resolutions and weights the output by the EAF. A semantic head (decoder) upsamples the output of this attention-based encoding and predicts the final output.}
  \label{fig:architecture}
\end{figure*}

Due to the coarse BEV grid, cells of the BEV feature map that are closer to the camera are farther apart from each other than the epipolar lines of cells that are more distant from the camera.
Our approach simplifies the modeling of epipolar geometry. The projection of a BEV cell to an epipolar line should be a projection to a line with a certain width, rather than an infinitesimally wide line.
For autonomous driving, we assume that the camera's principle axis is approximately parallel to the BEV plane. Thus, the projection of such a cell can be approximated by a line with a constant width. The width of this line depends on the relative distance of the BEV cell to the camera, the cell's size and the camera calibration. Furthermore, the approximation with constant width can also account for minor inaccuracies, such as those encountered when driving uphill.
For this reason, parameter $\lambda_{q,i}$ is used to scale the width of the Gaussian depending on the distance of the BEV cell corresponding to query $q$ and the origin of the $i$'th camera. We can adjust the width of the relative distance by the $\lambda$, which is either a pre-defined or a trainable parameter.
This principle is shown in \cref{fig:epipolar-fields-front-view} for various locations in the BEV~grid.

\subsection{Architecture}
\label{sec:arch}

Based on the idea of attention weighting and Epipolar Attention Fields (EAFs), we propose \name, which is a network architecture that can solely rely on the EAFs for spatial correspondence. The architecture of our network is depicted in \cref{fig:architecture}. 
For each image view, the image backbone outputs multi-scale image feature maps.

For each pair of BEV grid position and image feature map, we compute the EAF.
The Epipolar Transformer Encoder iteratively attends the BEV grid queries to the multi-scale feature maps and weights the attention output with the EAF. The cross-attention does not rely on a positional encoding since the EAF encodes the spatial relationship. Finally, a decoder network upsamples the BEV feature map to the final BEV grid resolution and outputs semantic masks. 

\section{Experiments}
For the experiments, we consider the tasks of map and vehicle segmentation. Map segmentation refers to predicting a semantic mask for the drivable area, while vehicle segmentation refers to segmenting the orthogonal projection of vehicles onto the ground plane.
In line with previous work~\cite{hu2021iccv, philion2020eccv, zhou2022cvpr}, our output BEV grid size is set to 200~$\times$~200 cells with a resolution of 0.5\,m per cell, thus corresponding to a 100 m $\times$ 100\,m area around the vehicle. The network performs a per timestep prediction without the utilisation of past frames.
The advantages of our primary contribution, the Epipolar Attention Fields, are comprehensively outlined in our ablation studies.

\subsection{Datasets}
Our experiments are performed on the autonomous driving datasets nuScenes~\cite{caesar2020cvpr} and Argoverse 2 (AV2)~\cite{wilson2021neurips}. 
nuScenes contains 1,000 scenes with multi-modal (camera, radar, LiDAR) measurements with 3D bounding box annotations and semantic maps. The scenes are captured in Boston and Singapore. Each scene is approximately 20 seconds long.
Similarly, Argoverse 2 (AV2)~\cite{wilson2021neurips} provides data for 1,000 sequences, each composed of multiple timesteps. It was recorded in six major cities in the US and contains thirty object categories.
For our experiments, we use the data from the six cameras for nuScenes and the data from the seven ring cameras for AV2.
For the vehicle segmentation task, we follow Zhou and Kr\"ahenb\"uhl~\cite{zhou2022cvpr} and project all 3D bounding boxes orthographically onto the ground for all instances that belong to the superclass vehicle. The ground truth is composed of a binary mask that segments the map into foreground (vehicle) and background. This mask is defined for the BEV area in ego vehicle coordinates. As done for most of the previous work on nuScenes, we only consider vehicles that have an appearance visibility~\textgreater{}40\% as annotated in the dataset.
Inherent to vision-based approaches is the correlation between the distance of a vehicle to the ego vehicle and the accuracy of its estimated location on the map. Furthermore, for the application of autonomous driving, we may have stronger requirements on the accuracy for near-distance vehicles than for vehicles that are farther away. We therefore also evaluate the vehicle segmentation task for different 10m-ranges of distance to the ego vehicle.

\begin{table}[tb]
  \caption{Comparisons with state-of-the-art methods on the nuScenes validation set.\ We report the mIoU for semantic segmentation of the drivable area (drivable) and the orthogonal vehicle projection (vehicle). LSS uses visibility $>0\%$, all others use visibility $>40\%$.\\ $\dagger$: Approach uses additional sensors, temporal information and/or a significantly larger backbone.}
  \label{tab:results_all}
  \centering
  \begin{tabular}{@{}lC{2cm}C{2cm}@{}}
    \toprule
     & Drivable & Vehicle\\
    \midrule
    LSS~\cite{philion2020eccv}              & 72.9 & 32.1 \\
    FIERY static ~\cite{hu2021iccv}        & -     & 35.8  \\
    CVT~\cite{zhou2022cvpr}                & 74.3  & 36.0  \\
    M$^2$BEV~\cite{xie2022arxiv}                & 77.2  & -     \\
    GKT~\cite{chen2022arxiv}            & -     & 38.0  \\
    BAEFormer~\cite{pan2023cvpr}       & 76.0  & 38.9  \\
    \textbf{\name\ (ours)}                  & \textbf{78.0} & \textbf{39.0} \\
  \midrule
  BEVFormer (S)~\cite{li2022eccv}$^\dagger$    & 80.7  & 43.2  \\
  BEVDepth~\cite{li2023aaai}$^\dagger$          & 82.7  & 45.0 \\        
  BEVFusion~\cite{liu2023icra}$^\dagger$       & 85.5  & - \\
  PETRv2~\cite{liu2023iccv}$^\dagger$             & 85.6  & 46.3 \\
  \bottomrule
  \end{tabular}
\end{table}

Additionally, we conduct cross-dataset experiments, \ie, we train the model on one dataset and evaluate it in a zero-shot transfer setting on data from another dataset. This means that the model has not seen any samples from the target dataset during training and therefore the model may be exposed to a domain shift during cross-dataset evaluation.
Furthermore, recent research \cite{yuan2024wacv, lilja2024cvpr} has shown that the locations of nuScenes driving scenes used for training and validation overlap by more than 80\%. This data leakage stands in conflict with the common practice of having disjoint sets of training and validation samples. To address this issue, the authors propose new splits with a set of non-overlapping locations. We conduct experiments with the data split proposed by Yuan~\etal\cite{yuan2024wacv}. 

\subsection{Implementation Details}
The input images are resized to 224 $\times$ 480 pixels in line with GKT~\cite{chen2022arxiv}. 
We maintain the image ratio by cropping the image from the top accordingly. Images taken by front camera of the AV2 dataset are in portrait mode and thus we do not preserve its original aspect ratio for resizing.
As image backbone, we follow CVT~\cite{zhou2022cvpr} and make use of EfficientNet-B4~\cite{tan2019icml}. 
By default, we use backbone image feature maps at scales of 1/4 and 1/16 for training as done by previous work~\cite{chen2022arxiv}.
We employ an Atrous Spatial Pyramid Pooling (ASPP)~\cite{chen2018pami} semantic segmentation head for the map and vehicle prediction. 
The training and evaluation are performed in the MMDetection3D~\cite{mmdet3d2020} framework and we selectively use pixel augmentation. Following CVT~\cite{zhou2022cvpr}, we use AdamW~\cite{kingma2015iclr} optimizer with one-cycle learning rate scheduling \cite{smith2019aiml} and employ focal loss~\cite{lin2017iccv-flfd} for the segmentation loss. The networks are trained on 4 GPUs with a batch size of 4. If not otherwise mentioned, we train for 30 epochs. The Intersection-over-Union (IoU) between the predicted and the ground truth semantic maps serves as the evaluation metric. The hyperparameter $\lambda$ controls the distance strength of the epipolar fields, \ie, the width of the distribution with respect to the distance to the ego vehicle. In \cref{sec:ablation}, we examine the influence of the choice of $\lambda$ and different feature map resolutions. For all other experiments, $\lambda$ is set to 1.0.

\subsection{Results}
\label{sec:results}

\cref{tab:results_all}  compares our model to prior work on the original dataset split. Our approach outperforms other camera-only methods by over 2\% mIoU for semantic segmentation of the drivable area. Our \name\ surpasses the original CVT by almost 4\%. 
For vehicle segmentation, our method performs on par with the recent work BAEFormer~\cite{pan2023cvpr}. 
In addition, we present results for other approaches~\cite{li2023aaai, li2022eccv, liu2023icra, liu2023iccv}, which leverage temporal and/or multi-modal data and employ significantly larger backbones and/or input resolution. 

\begin{table}[tb]
  \caption{Cross-dataset evaluation for AV2 and nuScenes. The experiments show the influence of changes in camera parameters for different camera setups in a zero-shot transfer setting. Methods are trained on the source dataset and then they are evaluated on the target dataset without retraining. We denote the zero-shot transfer from source dataset to target dataset as: $\text{source dataset}$ $\rightarrow \text{target dataset}$. AV2* is the AV2 dataset where the front-left stereo camera was used instead of the front ring camera.}
  \label{tab:cross-ds}
  \centering
  \begin{tabular}{C{1.9cm}|C{1.3cm}|C{1.8cm}C{1.8cm}@{}}
    \toprule
    & Training Epochs & CVT~\cite{zhou2022cvpr} & EAFormer (ours) \\
    \midrule
    \multirow{2}{1.9cm}{\centering{$\text{nuScenes}$} $\rightarrow \text{nuScenes}$} & 12 & 34.31 & 38.18\\
    & 30 & 36.69 & 38.98\\
    \midrule
    \multirow{2}{1.9cm}{\centering{$\text{AV2}$} \\ $\rightarrow \text{AV2}$} & 12 & 38.00 & 38.66 \\
    & 30 & 38.47 & 39.60 \\
    \midrule
    \multirow{2}{1.9cm}{\centering{$\text{AV2}$} \\ $\rightarrow \text{AV2*}$} & 12 & 30.92 & 32.99 \\
    & 30 & 31.21 & 33.91 \\
    \midrule
    \multirow{2}{1.9cm}{\centering{$\text{nuScenes}$} \\ $\rightarrow \text{AV2}$} & 12 & 7.78 & 19.72 \\
    & 30 & 7.86 & 14.00 \\
    \midrule
    \multirow{2}{1.9cm}{\centering{$\text{AV2}$} \\ $\rightarrow \text{nuScenes}$} & 12 & 3.07 & 12.17 \\
    & 30 & 2.70 & 11.44 \\
  \bottomrule
  \end{tabular}
\end{table}

\begin{figure*}[tb]
  \centering
  \includegraphics[width=0.89\linewidth]{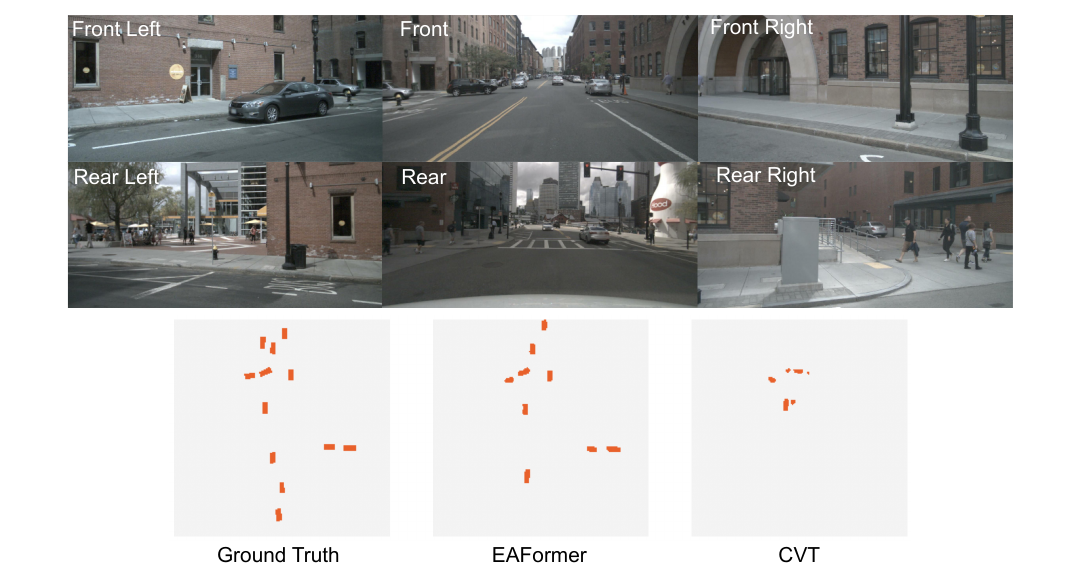}
  \caption{Visualization of zero-shot transfer performance of \name\ and CVT for vehicle segmentation. The models were trained on Argoverse 2 (AV2) and evaluated on nuScenes.}
  \label{fig:zero-shot}
\end{figure*}

\begin{table}[tb]
  \caption{Performance comparison for different splits. We report the mIoU for semantic segmentation of the drivable area for the original nuScenes data split and the disjoint split without data leakage~\cite{yuan2024wacv}. Both models were trained for 30 epochs.}
  \label{tab:new-split}
  \centering
  \begin{tabular}{@{}p{2cm}|C{2.3cm}C{2.7cm}@{}}
    \toprule
    & CVT~\cite{zhou2022cvpr} & EAFormer (ours) \\
    \midrule
    Original Split & 76.02 & 78.04 \\
    New Split & 54.23 & 58.06 \\
    \midrule
    Difference & -21.79 & \textbf{-19.98} \\
  \bottomrule
  \end{tabular}
\end{table}

In \cref{tab:cross-ds}, we present the cross-dataset evaluation for our method and compare it to a re-implementation of CVT with our settings to demonstrate the ability of the models to transfer to unseen camera setups. 
As previously mentioned, the network is trained for the specified number of epochs on the source dataset. Its performance is then evaluated in a zero-shot setting on the target dataset.
After training on AV2, CVT's ability to segment vehicles decreases significantly compared to its performance on the source dataset. This suggests that CVT does not adapt well to different camera setups because the positional encoding is learned for a fixed set of camera parameters.  However, our approach also experiences a considerable reduction in performance for the cross-dataset evaluation.
We attribute the performance difference partially to the domain shift of the data in the different datasets. When comparing nuScenes as the source dataset and AV2 as the target dataset, we also observe a substantial performance degradation of CVT, while EAFormer retains a stronger performance. 
In conclusion, our approach, \name, demonstrates approximately twice the performance for the transfer from nuScenes to AV2 and approximately four times the performance for the transfer from AV2 to nuScenes.
Qualitative results for the zero-transfer for CVT and \name\ are depicted in \cref{fig:zero-shot}.
To assess the model’s adaptability while avoiding domain shift between AV2 and nuScenes, we evaluated models trained on the AV2 dataset using a modified AV2* dataset. In this modified dataset, we replaced the front ring camera with the single front-left camera from the two stereo cameras. The result can be found in \cref{tab:cross-ds}. As in the previous case, EAFormer demonstrated superior performance compared to CVT.

\begin{table}[t]
  \caption{Ablation about architecture components. A reimplementation of CVT~\cite{zhou2022cvpr} serves as baseline. We report the performance for drivable area and vehicle segmentation for the original nuScenes split.}
  \label{tab:component-ablation}
  \centering
  \begin{tabular}{@{}p{4.5cm}C{1.4cm}C{1.4cm}@{}}
    \toprule
     &  Vehicle & Drivable\\ 
    \midrule
    Baseline                        & 36.69             & 76.02\\ 
     + ASPP                         & 37.39             & 77.14\\ 
     + epipolar attention weighting & 38.47             & 77.85\\ 
     -- pos. encoding               & \textbf{38.98}    & 78.04\\ 
     \midrule
     + different feature map scales & 38.19             & 78.89\\ 
     + learnable distance strength  & 38.25             & 79.28\\ 
     + pixel augmentation           & 38.37             & \textbf{79.54}\\ 
  \bottomrule
  \end{tabular}
\end{table}

\begin{table*}[tb]
  \caption{Distance-based evaluation (mIoU) for vehicle segmentation on nuScenes.}
  \label{tab:distance-based}
  \centering
  \begin{tabular}{L{2.7cm}|C{1.5cm}|C{1.5cm}C{1.5cm}C{1.5cm}C{1.5cm}C{1.5cm}|C{1.5cm}}
    \toprule
    & Epochs & 0 - 10 m & 10 - 20 m & 20 - 30 m & 30 - 40 m & 40 - 50 m & mIoU \\
    \midrule
    CVT & 12 & 67.28 & 53.01 & 37.42 & 22.60 & 11.02 & 34.02  \\
    EAFormer (ours) & 12 & 68.59 & 55.33 & 40.93 & 27.54 & 16.42 & 36.70 \\
    \midrule
    CVT & 30 & 68.62 & 55.38 & 40.37 & 26.85 & 15.44 & 36.69 \\
    EAFormer (ours) & 30 & 71.53 & 57.64 & 41.76 & 30.8 & 18.71 & 38.98 \\
  \bottomrule
  \end{tabular}
\end{table*}

We further compare the performance differences not only for map segmentation on the original nuScenes data split, but also report the performance for the new disjoint split.
\cref{tab:new-split} shows the results for the different splits.
The performance of both approaches is markedly diminished. However, the decline in EAFormer is less severe, thereby indicating superior generalization behavior. With these experiments, we highlight the issue of outputting overconfident predictions for occlusions, e.g., due to densely populated scenes.
The vehicle-segmentation performance of CVT and EAFormer for different distance-to-ego ranges is presented in \cref{tab:distance-based}. 
For both models, far-range perception is inherently more challenging as the mIoU decreases for increasing distance. However, EAFormer demonstrates better performance for far-range perception compared to CVT. 

In summary, our evaluation suggests that our method yields competitive results for BEV semantic segmentation for drivable area and vehicles using Epipolar Attention Fields. 
Moreover, it demonstrates better generalization to unseen data and is more adaptable to new camera settings.
Further experiments for AV2 can be found in the supplementary materials, where we additionally replaced the front-view camera with the stereo camera.



\subsection{Ablation studies}
\label{sec:ablation}

We employ CVT~\cite{zhou2022cvpr} as our baseline network to quantify the improvements for the individual elements of our approach. Given the slight difference in image size, we replicated better results for CVT that were reported in the original publication.

\begin{table}

    \caption{Ablations about the image feature map resolution scales.}
    \label{tab:ablation_resolution}
    \centering
    \begin{tabular}{@{}C{3.5cm}C{1.5cm}C{1.5cm}@{}}
        \toprule
        Res. Scale Factors& Vehicle & Drivable \\
        \midrule
        (1/8), (1/16)           & 37.62 & 76.98 \\
        (1/4), (1/16)           & 38.76 & 78.04 \\
        (1/4), (1/8), (1/16)    & 38.13 & 77.91 \\
        (1/8), (1/16), (1/32)   & 38.19 & 78.89 \\
      \bottomrule
      \end{tabular}
      \vspace{-0.1cm}
\end{table}

\cref{tab:component-ablation} shows the influence of different design decisions for both tasks. It shows that our main contribution, the use of Epipolar Attention Fields to weight attention, results in a significant performance boost.
Additionally, removing positional encodings from the training process further improves performance. The results demonstrate the benefits of explicitly modeling spatial relationships with Epipolar Attention Fields as part of our \name\ architecture for this application.
The performance for segmentation of the drivable area can be improved by selecting the backbone image feature maps of scales 1/8, 1/16, and 1/32. Additionally, learning the distance strength parameter during training and using pixel augmentation yield minor increases in performance for drivable map segmentation. 
Despite this improvement, we emphasize that these minor model changes may promote overfitting for map segmentation, as the performance for vehicle segmentation decreases slightly. 
Visualizations for the qualitative differences in segmentation performance of the same network trained on the two different splits are depicted in the supplementary materials.
The influence of the resolution scales of the image feature maps is further illustrated in \cref{tab:ablation_resolution}. Using more but smaller feature map scales does not significantly change the performance while using only smaller feature map scales degrades segmentation performance.

We also ablate about the distance strength parameter $\lambda$ in \cref{tab:ablation_ds}. Learning the distance strength leads to a minor performance gain, while it will degrade the performance for large values. If the distance strength is too small, the Epipolar Attention Field will not capture all relevant semantic content. If the value is too large, performance will degrade due to imprecise spatial correlation.

\begin{table}

    \caption{Ablations about the distance strength parameter $\lambda$.}
    \label{tab:ablation_ds}
    \centering
    \begin{tabular}{@{}C{3.5cm}C{1.5cm}C{1.5cm}@{}}
    \toprule
    Distance Strength $\lambda$ & Vehicle & Drivable \\
    \midrule
    1.0             & 38.76 & 78.04 \\
    1.4             & 38.00  &  77.65\\
    learnable       & 38.77 & 78.10 \\
    \bottomrule
    \end{tabular}
    \vspace{-0.0cm}
\end{table}

\section{Conclusion}
In this paper, we presented a novel approach to encode the spatial relationship between images and BEV grids with Epipolar Attention Fields. This allows us to explicitly model known geometric correspondences into a transformer-based BEV model, which we termed \name. 
The experimental evaluation indicates that weighting the cross-view attention with Epipolar Attention Fields outperforms methods that relate camera and BEV features with learned positional encodings. The results also support the claim that our method generalizes better, highlighting its capability to better transfer to entirely new camera setups with different extrinsics and intrinsics.
In future work, we anticipate that the integration of temporal information from sequences of camera images can be similarly realized through the imposition of geometric constraints on the location of cameras.


{\small
\bibliographystyle{ieee_fullname}
\bibliography{glorified,new}
}

\end{document}

%% file: stachnisslab-latex.tex
\usepackage{graphics}           
\usepackage{times}              
\usepackage{amsmath}            
\usepackage{amssymb}            
\usepackage{graphicx}
\usepackage{algorithm}
\usepackage[noend]{algpseudocode}
\usepackage{booktabs}
\usepackage{color}
\usepackage{listings}
\usepackage{subfiles}
\definecolor{instructioncolor}{rgb}{.5,.5,.5}

\usepackage[font=small]{caption}

\def\eqref#1{Eq.~(\ref{#1})}

\makeatletter
\usepackage{xspace}
\DeclareRobustCommand\onedot{\futurelet\@let@token\@onedot}
\def\@onedot{\ifx\@let@token.\else.\null\fi\xspace}
 
\def\ie{i.e\onedot}

\def\etal{{et al}\onedot}
\makeatother

\usepackage{array}
\newcolumntype{L}[1]{>{\raggedright\let\newline\\\arraybackslash\hspace{0pt}}m{#1}}
\newcolumntype{C}[1]{>{\centering\let\newline\\\arraybackslash\hspace{0pt}}m{#1}}
\newcolumntype{R}[1]{>{\raggedleft\let\newline\\\arraybackslash\hspace{0pt}}m{#1}}

%% file: witte2025wacv.bbl
\begin{thebibliography}{10}\itemsep=-1pt

\bibitem{caesar2020cvpr}
Holger Caesar, Varun Bankiti, Alex~H. Lang, Sourabh Vora, Venice~E. Liong, Qiang Xu, Anush Krishnan, Yu Pan, Giancarlo Baldan, and Oscar Beijbom.
\newblock {nuScenes: A Multimodal Dataset for Autonomous Driving}.
\newblock In {\em Proc.~of the IEEE/CVF Conf.~on Computer Vision and Pattern Recognition (CVPR)}, 2020.

\bibitem{chen2018pami}
Liang-Chieh Chen, George Papandreou, Iasonas Kokkinos, Kevin Murphy, and Alan~L. Yuille.
\newblock {DeepLab: Semantic Image Segmentation withDeep Convolutional Nets, Atrous Convolution,and Fully Connected CRFs}.
\newblock {\em IEEE Trans.~on Pattern Analysis and Machine Intelligence (TPAMI)}, 40(4):834--848, 2018.

\bibitem{chen2024wacv}
Qiuxiao Chen and Xiaojun Qi.
\newblock {Residual Graph Convolutional Network for Bird's-Eye-View Semantic Segmentation}.
\newblock In {\em Proc.~of the IEEE Winter Conf.~on Applications of Computer Vision (WACV)}, 2024.

\bibitem{chen2022arxiv}
Shaoyu Chen, Tianheng Cheng, Xinggang Wang, Wenming Meng, Qian Zhang, and Wenyu Liu.
\newblock {Efficient and robust 2d-to-bev representation learning via geometry-guided kernel transformer}.
\newblock {\em arXiv preprint}, arXiv:2206.04584, 2022.

\bibitem{mmdet3d2020}
MMDetection3D Contributors.
\newblock {MMDetection3D: OpenMMLab} next-generation platform for general {3D} object detection.
\newblock \url{https://github.com/open-mmlab/mmdetection3d}, 2020.

\bibitem{han2024ral}
Chunrui Han, Jinrong Yang, Jianjian Sun, Zheng Ge, Runpei Dong, Hongyu Zhou, Weixin Mao, Yuang Peng, and Xiangyu Zhang.
\newblock {Exploring recurrent long-term temporal fusion for multi-view 3d perception}.
\newblock {\em IEEE Robotics and Automation Letters (RA-L)}, 9(7):6544--6551, 2024.

\bibitem{harley2023icra}
Adam~W Harley, Zhaoyuan Fang, Jie Li, Rares Ambrus, and Katerina Fragkiadaki.
\newblock {Simple-bev: What really matters for multi-sensor bev perception?}
\newblock In {\em Proc.~of the IEEE Intl.~Conf.~on Robotics \& Automation (ICRA)}, 2023.

\bibitem{hartley2004book}
Richard Hartley and Andrew Zisserman.
\newblock {\em {Multiple View Geometry in Computer Vision}}.
\newblock Cambridge University Press, second edition, 2004.

\bibitem{he2020cvpr-epipolar}
Yihui He, Rui Yan, Katerina Fragkiadaki, and Shoou-I Yu.
\newblock {Epipolar transformers}.
\newblock In {\em Proc.~of the IEEE/CVF Conf.~on Computer Vision and Pattern Recognition (CVPR)}, 2020.

\bibitem{hu2021iccv}
Anthony Hu, Zak Murez, Nikhil Mohan, Sof{\'\i}a Dudas, Jeffrey Hawke, Vijay Badrinarayanan, Roberto Cipolla, and Alex Kendall.
\newblock {Fiery: Future instance prediction in bird's-eye view from surround monocular cameras}.
\newblock In {\em Proc.~of the IEEE/CVF Intl.~Conf.~on Computer Vision (ICCV)}, 2021.

\bibitem{huang2021arxiv-bevdet}
Junjie Huang, Guan Huang, Zheng Zhu, Yun Ye, and Dalong Du.
\newblock {Bevdet: High-performance multi-camera 3d object detection in bird-eye-view}.
\newblock {\em arXiv preprint}, arXiv:2112.11790, 2021.

\bibitem{kingma2015iclr}
Diederik~P. Kingma and Jimmy Ba.
\newblock {Adam: {A} Method for Stochastic Optimization}.
\newblock In {\em Proc.~of the Intl.~Conf.~on Learning Representations (ICLR)}, 2015.

\bibitem{li2023aaai}
Yinhao Li, Zheng Ge, Guanyi Yu, Jinrong Yang, Zengran Wang, Yukang Shi, Jianjian Sun, and Zeming Li.
\newblock {Bevdepth: Acquisition of reliable depth for multi-view 3d object detection}.
\newblock In {\em Proc.~of the Conf.~on Advancements of Artificial Intelligence (AAAI)}, 2023.

\bibitem{li2022eccv}
Zhiqi Li, Wenhai Wang, Hongyang Li, Enze Xie, Chonghao Sima, Tong Lu, Yu Qiao, and Jifeng Dai.
\newblock {Bevformer: Learning bird's-eye-view representation from multi-camera images via spatiotemporal transformers}.
\newblock In {\em Proc.~of the Europ.~Conf.~on Computer Vision (ECCV)}, 2022.

\bibitem{lilja2024cvpr}
Adam Lilja, Junsheng Fu, Erik Stenborg, and Lars Hammarstrand.
\newblock {Localization is all you evaluate: Data leakage in online mapping datasets and how to fix it}.
\newblock In {\em Proc.~of the IEEE/CVF Conf.~on Computer Vision and Pattern Recognition (CVPR)}, 2024.

\bibitem{lin2017iccv-flfd}
Tsung-Yi Lin, Priya Goyal, Ross Girshick, Kaiming He, and Piotr Doll{\'a}r.
\newblock {Focal Loss for Dense Object Detection}.
\newblock In {\em Proc.~of the IEEE Intl.~Conf.~on Computer Vision (ICCV)}, 2017.

\bibitem{liu2022eccv}
Yingfei Liu, Tiancai Wang, Xiangyu Zhang, and Jian Sun.
\newblock {Petr: Position embedding transformation for multi-view 3d object detection}.
\newblock In {\em Proc.~of the Europ.~Conf.~on Computer Vision (ECCV)}, 2022.

\bibitem{liu2023iccv}
Yingfei Liu, Junjie Yan, Fan Jia, Shuailin Li, Aqi Gao, Tiancai Wang, and Xiangyu Zhang.
\newblock {Petrv2: A unified framework for 3d perception from multi-camera images}.
\newblock In {\em Proc.~of the IEEE/CVF Intl.~Conf.~on Computer Vision (ICCV)}, 2023.

\bibitem{liu2023icra}
Zhijian Liu, Haotian Tang, Alexander Amini, Xinyu Yang, Huizi Mao, Daniela~L Rus, and Song Han.
\newblock {Bevfusion: Multi-task multi-sensor fusion with unified bird's-eye view representation}.
\newblock In {\em Proc.~of the IEEE Intl.~Conf.~on Robotics \& Automation (ICRA)}, 2023.

\bibitem{ma2021bmvc}
Haoyu Ma, Liangjian Chen, Deying Kong, Zhe Wang, Xingwei Liu, Hao Tang, Xiangyi Yan, Yusheng Xie, Shih-Yao Lin, and Xiaohui Xie.
\newblock {Transfusion: Cross-view fusion with transformer for 3d human pose estimation}.
\newblock In {\em Proc.~of British Machine Vision Conf. (BMVC)}, 2021.

\bibitem{pan2023cvpr}
Cong Pan, Yonghao He, Junran Peng, Qian Zhang, Wei Sui, and Zhaoxiang Zhang.
\newblock {BAEFormer: Bi-Directional and Early Interaction Transformers for Bird's Eye View Semantic Segmentation}.
\newblock In {\em Proc.~of the IEEE/CVF Conf.~on Computer Vision and Pattern Recognition (CVPR)}, 2023.

\bibitem{peng2023wacv}
Lang Peng, Zhirong Chen, Zhangjie Fu, Pengpeng Liang, and Erkang Cheng.
\newblock {BEVSegFormer: Bird's Eye View Semantic Segmentation From Arbitrary Camera Rigs}.
\newblock In {\em Proc.~of the IEEE Winter Conf.~on Applications of Computer Vision (WACV)}, 2023.

\bibitem{philion2020eccv}
Jonah Philion and Sanja Fidler.
\newblock {Lift, splat, shoot: Encoding images from arbitrary camera rigs by implicitly unprojecting to 3d}.
\newblock In {\em Proc.~of the Europ.~Conf.~on Computer Vision (ECCV)}, 2020.

\bibitem{qin2023cvpr}
Zequn Qin, Jingyu Chen, Chao Chen, Xiaozhi Chen, and Xi Li.
\newblock {UniFusion: Unified Multi-View Fusion Transformer for Spatial-Temporal Representation in Bird's-Eye-View}.
\newblock In {\em Proc.~of the IEEE/CVF Conf.~on Computer Vision and Pattern Recognition (CVPR)}, 2023.

\bibitem{roddick2018arxiv}
Thomas Roddick, Alex Kendall, and Roberto Cipolla.
\newblock {Orthographic feature transform for monocular 3d object detection}.
\newblock {\em arXiv preprint}, arXiv:1811.08188, 2018.

\bibitem{saha2022icra}
Avishkar Saha, Oscar Mendez, Chris Russell, and Richard Bowden.
\newblock {Translating images into maps}.
\newblock In {\em Proc.~of the IEEE Intl.~Conf.~on Robotics \& Automation (ICRA)}, 2022.

\bibitem{salazar2023arxiv}
Gustavo Salazar-Gomez, Wenqian Liu, Manuel Diaz-Zapata, David Sierra-Gonzalez, and Christian Laugier.
\newblock {TLCFuse: Temporal Multi-Modality Fusion Towards Occlusion-Aware Semantic Segmentation-Aided Motion Planning}.
\newblock {\em arXiv preprint}, arXiv:2311.05319, 2023.

\bibitem{smith2019aiml}
Leslie~N Smith and Nicholay Topin.
\newblock Super-convergence: Very fast training of neural networks using large learning rates.
\newblock {\em Artificial Intelligence and Machine Learning for Multi-Domain Operations Applications}, 11006:369--386, 2019.

\bibitem{tan2019icml}
Mingxing Tan and Quoc Le.
\newblock {Efficientnet: Rethinking model scaling for convolutional neural networks}.
\newblock In {\em Proc.~of the Intl.~Conf.~on Machine Learning (ICML)}, 2019.

\bibitem{vaswani2017neurips}
Ashish Vaswani, Noam Shazeer, Niki Parmar, Jakob Uszkoreit, Llion Jones, Aidan~N. Gomez, Lukasz Kaiser, and Illia Polosukhin.
\newblock {Attention Is All You Need}.
\newblock In {\em Proc.~of the Conf.~on Neural Information Processing Systems (NeurIPS)}, 2017.

\bibitem{wang2023iccv}
Haiyang Wang, Hao Tang, Shaoshuai Shi, Aoxue Li, Zhenguo Li, Bernt Schiele, and Liwei Wang.
\newblock {UniTR: A Unified and Efficient Multi-Modal Transformer for Bird's-Eye-View Representation}.
\newblock In {\em Proc.~of the IEEE/CVF Intl.~Conf.~on Computer Vision (ICCV)}, 2023.

\bibitem{wilson2021neurips}
Benjamin Wilson, William Qi, Tanmay Agarwal, John Lambert, Jagjeet Singh, Siddhesh Khandelwal, Bowen Pan, Ratnesh Kumar, Andrew Hartnett, Jhony~Kaesemodel Pontes, Deva Ramanan, Peter Carr, and James Hays.
\newblock {Argoverse 2: Next Generation Datasets for Self-driving Perception and Forecasting}.
\newblock In {\em Proc.~of the Conf.~on Neural Information Processing Systems (NeurIPS)}, 2021.

\bibitem{xie2022arxiv}
Enze Xie, Zhiding Yu, Daquan Zhou, Jonah Philion, Anima Anandkumar, Sanja Fidler, Ping Luo, and Jose~M Alvarez.
\newblock {M$^2$BEV: Multi-Camera Joint 3D Detection and Segmentation with Unified Birds-Eye View Representation}.
\newblock {\em arXiv preprint}, arXiv:2204.05088, 2022.

\bibitem{xu2022arxiv}
Runsheng Xu, Zhengzhong Tu, Hao Xiang, Wei Shao, Bolei Zhou, and Jiaqi Ma.
\newblock {CoBEVT: Cooperative bird's eye view semantic segmentation with sparse transformers}.
\newblock {\em arXiv preprint}, arXiv:2207.02202, 2022.

\bibitem{yan2023iccv}
Junjie Yan, Yingfei Liu, Jianjian Sun, Fan Jia, Shuailin Li, Tiancai Wang, and Xiangyu Zhang.
\newblock (cross modal transformer: Towards fast and robust 3d object detection).
\newblock In {\em Proc.~of the IEEE/CVF Intl.~Conf.~on Computer Vision (ICCV)}, 2023.

\bibitem{yang2023cvpr}
Chenyu Yang, Yuntao Chen, Hao Tian, Chenxin Tao, Xizhou Zhu, Zhaoxiang Zhang, Gao Huang, Hongyang Li, Yu Qiao, Lewei Lu, et~al.
\newblock {BEVFormer v2: Adapting Modern Image Backbones to Bird's-Eye-View Recognition via Perspective Supervision}.
\newblock In {\em Proc.~of the IEEE/CVF Conf.~on Computer Vision and Pattern Recognition (CVPR)}, 2023.

\bibitem{yang2023iccv}
Jiayu Yang, Enze Xie, Miaomiao Liu, and Jose~M Alvarez.
\newblock {Parametric Depth Based Feature Representation Learning for Object Detection and Segmentation in Bird's-Eye View}.
\newblock In {\em Proc.~of the IEEE/CVF Intl.~Conf.~on Computer Vision (ICCV)}, 2023.

\bibitem{yuan2024wacv}
Tianyuan Yuan, Yicheng Liu, Yue Wang, Yilun Wang, and Hang Zhao.
\newblock {Streammapnet: Streaming mapping network for vectorized online hd map construction}.
\newblock In {\em Proc.~of the IEEE Winter Conf.~on Applications of Computer Vision (WACV)}, 2024.

\bibitem{zhang2022arxiv-beverse}
Yunpeng Zhang, Zheng Zhu, Wenzhao Zheng, Junjie Huang, Guan Huang, Jie Zhou, and Jiwen Lu.
\newblock {Beverse: Unified perception and prediction in birds-eye-view for vision-centric autonomous driving}.
\newblock {\em arXiv preprint}, arXiv:2205.09743, 2022.

\bibitem{zhou2022cvpr}
Brady Zhou and Philipp Kr{\"a}henb{\"u}hl.
\newblock {Cross-view transformers for real-time map-view semantic segmentation}.
\newblock In {\em Proc.~of the IEEE/CVF Conf.~on Computer Vision and Pattern Recognition (CVPR)}, 2022.

\bibitem{zhou2023cvpr}
Hongyu Zhou, Zheng Ge, Zeming Li, and Xiangyu Zhang.
\newblock {Matrixvt: Efficient multi-camera to bev transformation for 3d perception}.
\newblock In {\em Proc.~of the IEEE/CVF Intl.~Conf.~on Computer Vision (ICCV)}, 2023.

\bibitem{zou2023arxiv}
Jian Zou, Tianyu Huang, Guanglei Yang, Zhenhua Guo, and Wangmeng Zuo.
\newblock {UniM$^2$AE: Multi-modal Masked Autoencoders with Unified 3D Representation for 3D Perception in Autonomous Driving}.
\newblock {\em arXiv preprint}, arXiv:2308.10421, 2023.

\end{thebibliography}
